\documentclass[conference]{IEEEtran}
\IEEEoverridecommandlockouts
\usepackage{booktabs} % For formal tables
\usepackage{multirow}
\usepackage{color}
\usepackage{subfigure} % Written by Steven Douglas Cochran
\usepackage{array}
\usepackage{breqn}
\usepackage{{inputenc}}
\usepackage{balance}
\usepackage{multirow,tabularx}
\usepackage{longtable}
\usepackage{booktabs,longtable}
\usepackage{hhline}
\usepackage[flushleft]{threeparttable}
\usepackage{algorithm}
\usepackage{algorithmic}
\usepackage{epsfig}
\usepackage{graphicx}
\usepackage{epstopdf}
\usepackage{thmtools}
\usepackage{enumitem}
\usepackage{verbatim}
\usepackage{amsfonts}
\usepackage{hyperref}
\hypersetup{colorlinks=false,linkcolor=blue,urlcolor=blue,citecolor=red}

\usepackage{cite}
\usepackage{amsmath,amssymb,amsfonts}
\usepackage{textcomp}
\usepackage{xcolor}
\usepackage{caption}
\usepackage{dsfont}

\usepackage{xcolor}
\newcommand{\linebreakand}{%
  \end{@IEEEauthorhalign}
  \hfill\mbox{}\par
  \mbox{}\hfill\begin{@IEEEauthorhalign}
}

\def\BibTeX{{\rm B\kern-.05em{\sc i\kern-.025em b}\kern-.08em
    T\kern-.1667em\lower.7ex\hbox{E}\kern-.125emX}}
\begin{document}

\title{Explainable Sparse Knowledge Graph Completion via High-order Graph Reasoning Network}

\author{\IEEEauthorblockN{Weijian Chen~$^{1}$, Yixin Cao~$^{2}$, Fuli Feng~$^{1}$, Xiangnan He~$^{1}$, Yongdong Zhang~$^{1}$}
\IEEEauthorblockA{$^{1}$\textit{University of Science and Technology of China,} Hefei, China \\
naure@mail.ustc.edu.cn, \{fulifeng93,xiangnanhe\}@gmail.com, zhyd73@ustc.edu.cn \\
~$^{2}$\textit{Singapore Management University,}
Singapore \\
caoyixin2011@gmail.com}
}

\maketitle

\begin{abstract}

Knowledge Graphs (KGs) are becoming increasingly essential infrastructures in many applications while suffering from incompleteness issues. The KG completion task (KGC) automatically predicts missing facts based on an incomplete KG. However, existing methods perform unsatisfactorily in real-world scenarios. On the one hand, their performance will dramatically degrade along with the increasing sparsity of KGs. On the other hand, the inference procedure for prediction is an untrustworthy black box.

This paper proposes a novel explainable model for sparse KGC, compositing high-order reasoning into a graph convolutional network, namely HoGRN. It can not only improve the generalization ability to mitigate the information insufficiency issue but also provide interpretability while maintaining the model's effectiveness and efficiency. There are two main components that are seamlessly integrated for joint optimization. First, the high-order reasoning component learns high-quality relation representations by capturing endogenous correlation among relations. This can reflect logical rules to justify a broader of missing facts. Second, the entity updating component leverages a weight-free Graph Convolutional Network (GCN) to efficiently model KG structures with interpretability. Unlike conventional methods, we conduct entity aggregation and design composition-based attention in the relational space without additional parameters. The lightweight design makes HoGRN better suitable for sparse settings. For evaluation, we have conducted extensive experiments—the results of HoGRN on several sparse KGs present impressive improvements (9\% MRR gain on average). Further ablation and case studies demonstrate the effectiveness of the main components. Our codes will be released upon acceptance.

\end{abstract}


\section{Introduction}
\label{sec:introduction}

% This is an introduction.

\begin{figure}[t]
 \centering
 \includegraphics[width=0.99\columnwidth]{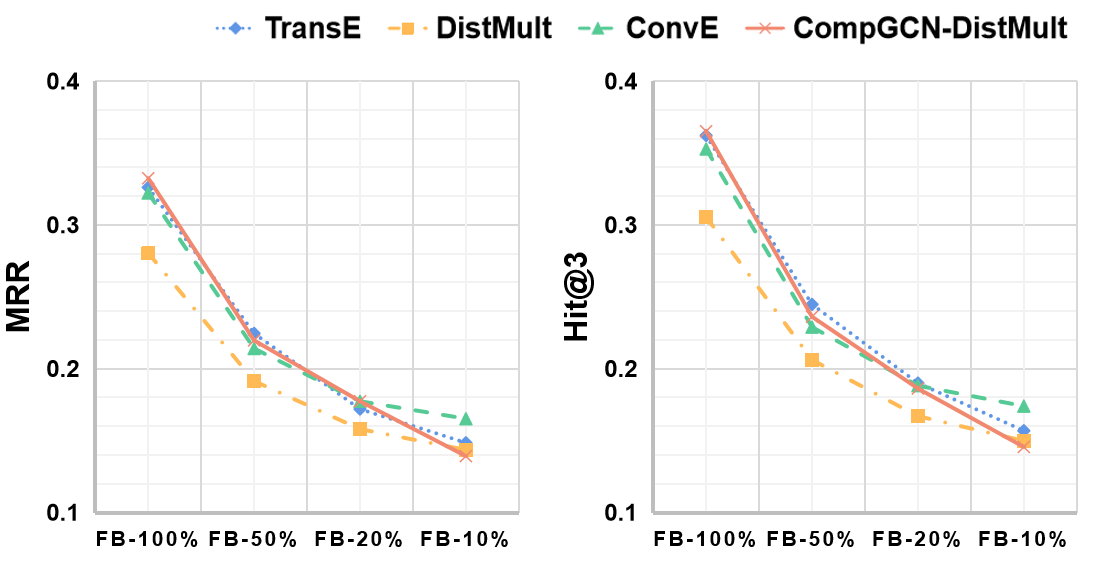}
%   \vspace{-10pt}
    \caption{
    Link prediction results of different KGC models on FB15K-237 (short for FB) and its sparse subsets (50\%, 20\%, and 10\% denotes sparsity degree, detailed statistics can be found in Section~\ref{sec:experiments}).
    }
%  \vspace{-10pt}
 \label{fig:sparse}
\end{figure}

\begin{figure*}[t]
 \centering
 \includegraphics[width=0.89\textwidth]{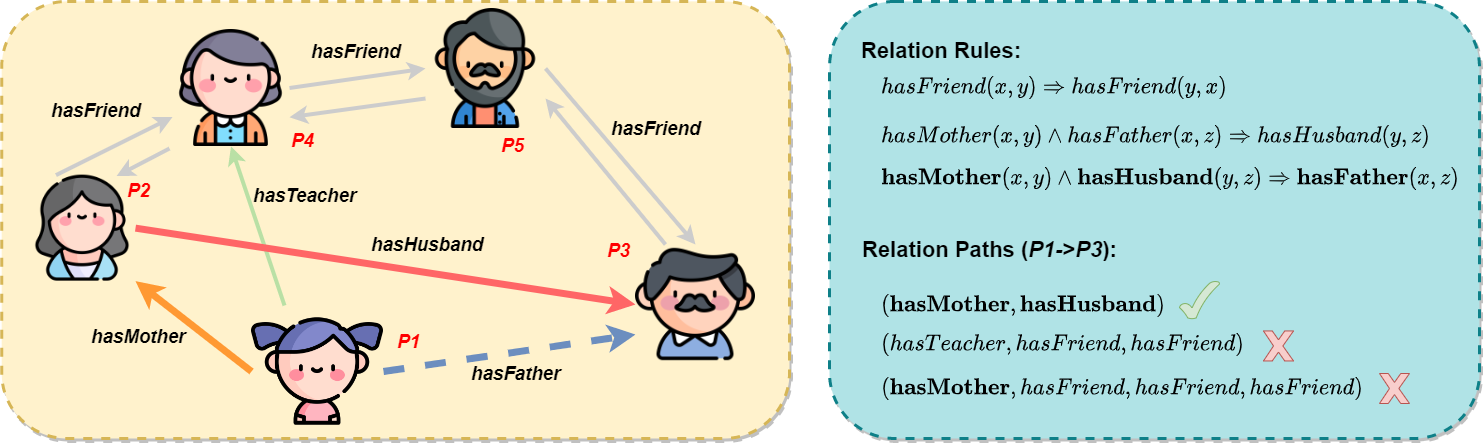}
%   \vspace{-10pt}
    \caption{
    Illustration of high-order reasoning among relations.
    On the left, the dotted line refers to the missing fact to be predicted, and the thickness of edges reflect their importance for the prediction. On the right, we present the reasoning rules behind the example and potential reasoning paths between entities $P1$ and $P3$. 
    }
%  \vspace{-10pt}
 \label{fig:toy}
\end{figure*}

Knowledge graphs (KGs) are one of the most effective ways to organize world facts in the form of a directed graph, where nodes denote entities and edges denote their relations. 
Recently, a series of KGs have been curated in various domains, including medicine \cite{DDI}, health care \cite{PDD}, and finance \cite{fintech}.
They are playing an increasingly important role in a variety of applications, such as drug discovery~\cite{KGNN}, user modeling~\cite{KGCN,KGAT}, dialog system~\cite{wu2021knowledge}, and question answering~\cite{bordes2014question,LEGO}.
However, existing KGs suffer from serious incompleteness issues. 
Due to the high cost of manual labeling, KG completion (KGC) becomes an essential task for predicting missing facts based on an incomplete KG.

Although existing KGC methods have achieved great success, we argue that their performance will dramatically degrade in real-world scenarios --- those KGs are more sparse than common KGC datasets.
As shown in Figure~\ref{fig:sparse}, our pilot study shows a downward trend in KGC performance along with the increasing data sparsity. 
The curves denote four state-of-the-art embedding-based models: TransE \cite{TransE}, DistMult \cite{DistMult}, ConvE \cite{ConvE}, and CompGCN \cite{CompGCN}. The x-axis denotes that we gradually remove links from a widely used dataset, FB15K-237~\cite{FB15K237}, where the resulting out-degree of entities on average are 22.32 (FB-100\%), 13.04 (FB-50\%), 7.53 (FB-20\%) and 5.84 (FB-10\%) on four subsets, respectively. \footnote{We follow the sparsity measurement of  DacKGR~\cite{DacKGR}, a lower average out-degree of nodes denotes a higher sparsity of the KG.} 
To alleviate the sparsity issue, DacKGR~\cite{DacKGR} improves another line of explainable KGC models that leverage reinforcement learning to discover multi-hop reasoning paths for each missing edge. In specific, they first pre-train an embedding-based KGC model and rely on it to guide the path selection.
However, the additional interpretability typically results in the loss of performance compared with embedding-based models and usually suffers from the inefficiency issue.

In this paper, we propose a novel KGC model that composites \textbf{H}igh-\textbf{o}rder reasoning into \textbf{GR}aph convolutional \textbf{N}etwork, namely \textbf{HoGRN}. 
It targets real scenarios where KGs are usually very sparse, and the completion process needs to be transparent for trustworthiness. 
To achieve the target, we highlight the following questions:
\begin{itemize}
    \item How to mitigate the performance drop caused by information insufficiency?
    \item How to provide interpretability while maintaining the model's effectiveness and efficiency?
\end{itemize}

To deal with the first challenge, we design a high-order reasoning component among relations to improve the generalization ability, so that HoGRN is expected to learn logic rules that is able to justify a broader of missing triples. 
As shown in Figure~\ref{fig:toy}, given any entities $x,y,z$ (we take $P1$, $P2$, $P3$ as an example here), if they hold $P1$ $hasMother$ $P2$ and $P2$ $hasHusband$ $P3$, it is more probable that $P1$ $hasFathe$r $P3$. 
Clearly, such rule suggests an endogenous correlation of relations ($hasMother,\ hasHusband$) to predict the missing edge $hasFather$, compared with other paths between $P1$ and $P3$, e.g., ($hasTeacher,\ hasFriend, \ hasFriend$).
To capture their endogenous correlations, we perform information flow among relations without introducing prior rule knowledge, which relies heavily on massive KG facts for rule mining and is unsuitable for sparse settings.
In specific, there are two modules: inter-relation learning to composite different relations along the same dimension, and intra-relation learning for self-enhancement by mixing its own dimensions.
Furthermore, we design two strategies for better optimization. 
Specifically, we first randomly mask some relations to encourage reconstruction by fusing information from related relations, since most relations are irrelevant and may be noisy in inter-relation learning. 
Second, we pose a relational contrastive constraint to avoid over-smoothing --- all relation vectors may become too similar to distinguish during iterative information fusion.

To address the second challenge, we leverage Graph Convolutional Network (GCN) with attention, rather than reinforcement learning, to efficiently model KG structures with interpretability. 
Unlike conventional GCN-based KGC methods, we leverage a weight-free GCN to update entity representations considering sparsity. 
On the one hand, the aggregation of neighbor entities doesn't require any parameters. 
In addition, we compute triple-specific attention scores for the weighted summation of their embeddings, and the attention is based on the relevance between head and tail entities with respect to the relation. 
On the other hand, we project all entities into the relational space for updating without additional parameters either. 
This is mainly because the same entity expresses different aspects of semantics conditioned on specific relations, and it is critical to distinguish them in GCN-based KGC models.
Specifically, we apply diagonalization constraints to the relation representation (learned in the above reasoning component) and leverage it as the projection matrix.

We not only seamlessly integrate the above two components --- high-order reasoning enhances relation representations, and relation defines entity aggregation as well as attention --- but also attention scores over edges reflect the reasoning paths. 
Through different paths between two entities, higher attentions suggest more information flow for the missing link prediction. 
Besides, HoGRN is a lightweight framework that removes unnecessary network weights and enjoys an efficient interpretable computation in sparse KG scenarios.

Our contributions can be summarized as follows:
\begin{itemize}
    \item We propose to capture high-order reasoning among relations to improve the KGC model generalization ability without using any prior rule guidance.
    \item We propose a novel sparse KGC model, HoGRN, that targets real scenarios considering the information insufficiency and interpretability without the loss of effectiveness and efficiency. 
    \item We have conducted extensive experiments on three sparse KGs and another two widely used KGC datasets. The results present impressive improvements (9\% MRR gain on average) of HoGRN in sparse settings and a comparable performance when more knowledge is available. Further ablation study and case study demonstrate the effectiveness of our main components.
\end{itemize}

\section{Related Work}
\label{sec:related}

% This is a related work.

Knowledge graph completion (KGC) aims at using existing information in KGs to complete the inference of unknown facts, which can be roughly divided into embedding-based methods and multi-hop reasoning-based methods. 
The basic idea of embedding-based methods is to learn representations of entities and relations to model the semantic association among triples (head entity, relation, tail entity). Then, they evaluate the plausibility of unknown facts through well-designed scoring functions. 
According to different design criteria of the scoring function, these approaches can be divided into translation-based \cite{TransE,TransH,TransR,RotatE}, semantic matching-based \cite{SME,RESCAL,DistMult,HOLE,TuckER,ComplEx}, and neural network-based \cite{dong2014knowledge,NTN,ConvE,ConvR,ConvKB,HypER,LTE-KGE}.
These methods are trained in an end-to-end manner while balancing efficiency and performance.
Different from the aforementioned embedding-based methods, multi-hop reasoning methods sacrifice some performance to improve interpretability.
They discover possible inference paths to complete the prediction of missing facts, which are especially necessary for many scenarios. 
Multi-hop reasoning methods generally include reinforcement learning-based and neural symbolic rules-based models. 
The former uses reinforcement learning algorithms that take entities/relations as states/actions and discover the best paths between target head and tail entities, including MINERVA \cite{MINERVA}, MultiHopKG \cite{MultiHopKG}, while the latter mines logical rules to help establish inference paths, such as NTP \cite{NTP}, NeurlLP \cite{NeuralLP}.

To improve the embedding-based KGC method, graph convolutional networks (GCNs) have been introduced to model the global structures~\cite{KGCN,KGAT,cao2019unifying,GENI}. 
GCN~\cite{GCN,GAT,CatGCN} is an efficient framework that can simultaneously integrate node representation and network structure and has gained wide attention with abundant applications, such as natural language processing \cite{yao2019graph}, user profiling \cite{rahimi2018semi,HGAT}, and recommendation systems \cite{PinSAGE,LightGCN,KGIN}. 
Most of the existing GCNs usually follow the idea of message passing neural networks \cite{MPNN} to complete the update of node representation through propagating information on the graph.
GCN-based KGC models usually use GCNs as the encoder to complete the representation learning of entities and relations. Then, they use embedding-based scoring functions as the decoder to evaluate the plausibility of facts \cite{RGCN}. 
This strategy further improves the performance of embedding-based methods because of the rich structural information contained in the refined representations. 
However, most of these approaches focus on entity updating \cite{DGCN,RGCN,WGCN,GraIL}, and those that consider relation updating are limited to shallow relation-entity interactions \cite{CompGCN,KEGNN,PathCON}. Our proposed HoGRN instead learns endogenous correlation among relations to enhance reasoning ability.

Despite the encouraging progress, most KGC approaches assume that there are sufficient triples to learn generic vectors of entities and relations, while little consideration is given to learning in sparse KG scenarios. This is rather common in practice, where severe information loss may lead to catastrophic degradation of model discrimination.
To solve the above problem, DacKGR \cite{DacKGR} proposed to guide the selection of reasoning paths for multi-hop reasoning KGC methods. They use a pre-trained embedding-based KGC model to screen each potential fact in the paths.
Although this compromise strategy gains interpretability, it does not perform as well as the embedding-based approaches (i.e., it actually degrades the performance of the pre-trained embedding-based KGC methods). 
In contrast, our proposed HoGRN follows the paradigm of the GCN-based KGC framework, which enhances the representation learning capability by integrating the weight-free GCNs.
On the one hand, it provides interpretability by introducing the attention mechanism on relation paths. On the other hand, it strengthens the quality of relation representation and alleviates sparsity through endogenous relational learning.
In the end, HoGRN achieves better performance than the embedding-based KGC models on sparse KGs while providing some interpretability for the predicted results.

% \section{Problem Formulation}
% \label{sec:formulation}

\section{Preliminaries}

\begin{table}[]
	\caption{Terms and notations used in the paper.}
	\label{tab:terms}
	\resizebox{0.48\textwidth}{!}{
		\begin{tabular}{c l}
			\hline
			Symbol  & Definition \\ \hline 
			$\mathcal{KG}$ & A knowledge graph \\
			$\mathcal{V}$ & The entity set with size $N$ \\ 
			$\mathcal{R}$ & The relation set with size $M$ \\  
			$\mathcal{T, T'}$ & A triplet set and its extension \\
			$(s, r, t)$ & One fact with head entity $s$, relation $r$, and tail entity $t$ \\
			$\mathcal{N}_t$ & Neighbor relation-entity pairs of the tail entity $t$ \\
			$\mathcal{P}_{s \rightarrow t}$ & The path set between entities $s$ and $t$ \\
			$\mathbf{e}_s, \mathbf{e}_r, \mathbf{e}_t$ & Embedding of entity and relation with dimensions $d$ \\
			$\mathbf{h}^l_s, \mathbf{z}^l_r, \mathbf{h}^l_t$ & Entity and relation hidden states after $l$-th GCN layers \\
			$\alpha_{srt}$ & Attention score between entities $s$ and $t$ under relation $r$ \\
			$\phi(\cdot)$ & Composition operation on entity and relation \\
			$diag(\cdot)$ & Vector diagonalization \\
			$concat(\cdot)$ & Vectors concatenation \\
			$vec(\cdot)$ & Matrix vectorization \\
			$\texttt {mask}(\cdot)$ & The stochastic relational masking operation \\
			$\mathcal{L}_{kge}$ & The principal loss calculated based on the scoring function  \\
			$\mathcal{L}_{rel}$ & The auxiliary relational contrastive loss \\
			\hline
		\end{tabular}
	}
\end{table}

\begin{table*}[ht]
    \centering
    \caption{Comparison of computational strategies for several representative GCN-based KGC models.
    Here, $\mathbf{W}^l$ is the linear transformation matrix for entity updating in $l$-th GCN layer, while $\mathbf{W}_r^l$ and $\mathbf{W}_{dir}^l$ specify the relation and direction, respectively.
    $\mathbf{W}_{rel}^l$ is the relational linear transformation matrix.
    $\alpha _r^l \in \mathbb{R}$ is a relation-specific learnable parameter.
    $\phi: \mathbb{R}^d \times \mathbb{R}^d \rightarrow \mathbb{R}^d$ is the composition operator, which can be element-wise subtraction, multiplication, or circular-correlation, etc.
    Note that since the triplets in KG were previously extended (see formula \ref{eq:trp_ext}), $\mathcal{N}_t$ already contains the incoming, outgoing, and self-information for the entity to be updated.
    For clarity, normalized coefficients and nonlinear activation functions are ignored here.}
    % }
    \label{tab:kgcn}
    % \vspace{-1.5mm}
    \begin{tabular}{cccc}
    \toprule
    & \textbf{Entity Updating} & \textbf{Relation Updating}\\
    \midrule
    RGCN \cite{RGCN}  & 
    $\mathbf{h}^{l+1}_t = f_{ent}\left(
    \left\{\mathbf{h}^l_s|(s,r)\in \mathcal{N}_t
    \right\}
    \right)  = \sum_{(s,r)\in \mathcal{N}_t} \mathbf{W}_r^l \mathbf{h}^l_s$ &
    $\mathbf{z}^{l+1}_r = f_{rel} \left (\mathbf{z}^l_r \right ) = \mathbf{z}^{l}_r$ \\
    WGCN \cite{WGCN} & 
    $\mathbf{h}^{l+1}_t = f_{ent}\left(
    \left\{\mathbf{h}^l_s|(s,r)\in \mathcal{N}_t
    \right\}
    \right)  = \sum_{(s,r)\in \mathcal{N}_t} \mathbf{W}^l (\alpha _{r}^l\mathbf{h}^l_s)$ &
    $\mathbf{z}^{l+1}_r = f_{rel} \left (\mathbf{z}^l_r \right ) = \mathbf{z}^{l}_r$ \\
    CompGCN \cite{CompGCN} & 
    $\mathbf{h}^{l+1}_t = f_{ent}\left(
    \left\{(\mathbf{h}^l_s,\mathbf{z}^l_r)|(s,r)\in \mathcal{N}_t
    \right\}
    \right)  = \sum_{(s,r)\in \mathcal{N}_t} \mathbf{W}_{dir}^l \phi(\mathbf{h}^l_s,\mathbf{z}^l_r)$ & $\mathbf{z}^{l+1}_r = f_{rel} \left (\mathbf{z}^l_r \right ) = \mathbf{W}_{rel}^l \mathbf{z}^{l}_r$ \\
    \midrule
    \multirow{2}{*}{HoGRN} & 
    $\mathbf{h}^{l+1}_t = f_{ent}\left(
    \left\{(\mathbf{h}^l_s,\mathbf{z}^l_r,\mathbf{h}^l_t)|(s,r)\in \mathcal{N}_t
    \right\}
    \right)  = \sum_{(s,r)\in \mathcal{N}_t} \alpha _{srt} \phi(\mathbf{h}^l_s,\mathbf{z}^l_r),$ & \multirow{2}{*}{$\mathbf{z}^{l+1}_r = f_{rel} \left (\left \{\mathbf{z}^l_k|k \in \mathcal{R} \right \} \right )$} \\
      & $\alpha _{srt}=tanh \left [\phi ^T(\mathbf{h}^l_s,\mathbf{z}^l_r)\phi (\mathbf{h}^l_t,\mathbf{z}^l_r) \right ]$ &  \\
    \bottomrule
    \end{tabular}
\end{table*}

A \textbf{Knowledge Graph (KG)} is typically formulated as a directed graph $\mathcal{KG}=\{\mathcal{V}, \mathcal{R}, \mathcal{T}\}$, where $\mathcal{V}$ is the entity set ($|\mathcal{V}|=N$), $\mathcal{R}$ is the relation set ($|\mathcal{R}|=M$), and $\mathcal{T}\subseteq \mathcal{V} \times \mathcal{R} \times \mathcal{V} $ is a set of triples. Each triple $(s,r,t)\in \mathcal{T}$ denotes a world fact that a head entity $s$ has relation $r$ with tail entity $t$. 

The task of \textbf{Knowledge Graph Completion (KGC)} is to predict missing/unknown facts based on existing knowledge in $\mathcal{KG}$ (i.e., $\mathcal{T}$). That is, KGC is to predict a tail entity $t$ from $\mathcal{V}$ that is the most compatible within an incomplete triple $(s,r,?)$, or to predict a head entity $s$ for $(?, r, t)$. 
This is also known as link prediction, the widely used KGC evaluation task currently.

Following previous work \cite{DacKGR}, the \textbf{sparsity} of KGs is measured by the average out-degree of entities. This reflects how many neighbor nodes are available on average for each entity in the KG. Note that a higher degree or more neighbors denote a low sparsity.

As convention, we denote the \textbf{embedding} vectors using bold fonts, such as the embedding of head entity $s$ as $\mathbf{e}_s \in \mathbb{R}^d$, the embedding of relation $r$ as $\mathbf{e}_r \in \mathbb{R}^d$, and the embedding of tail entity $t$ as $\mathbf{e}_t \in \mathbb{R}^d$. Besides, we consider the direction of each relation, i.e., we extend the triplet set as follows,
\begin{equation}
% \small
\mathcal{T}' = \mathcal{T} \cup \{(t,r^{-1},s) |(s,r,t) \in \mathcal{T}\} \cup \{(s,r_s,s) |s \in \mathcal{V}\}
\label{eq:trp_ext}
\end{equation}
where $r^{-1}$ is the inverse relation of $r$, and $r_s$ indicates a special self-loop relation connecting the entity itself. We summarize the terms and notations used in this paper in Table \ref{tab:terms}.

\section{Methodology}
\label{sec:methodology}

The goal of HoGRN is to alleviate the negative impacts of sparsity for KGC by considering real-world scenarios while ensuring efficient computation and interpretable results. In this section, we first introduce the overall framework, followed by each principal component. Then, we describe the training regime involving various score functions. Finally, we give a discussion compared with other GCN-based KGC models.

\subsection{Framework}

As shown in Figure~\ref{fig:framework}, there are two main components: Entity Updating and High-order Reasoning, which are jointly optimized. 1) To efficiently model KG structures with interpretability, entity updating learns entity embeddings using GCN with attention. Note that we adopt a weight-free fashion in aggregating entity features to minimize the learning burden of insufficient information. And, the aggregation considers the importance of neighbor entities conditioned on their relations via a composition-based attention mechanism. 2) The component of high-order reasoning targets enhancing relation embedding via two sub-modules: inter-relation learning to capture endogenous correlation among relations (i.e., logic rules) for generalization ability, and intra-relation learning that refines the relation itself through the mix of different vector dimensions. Besides, stochastic relational masking and relational contrastive constraints are designed for better optimization --- fusing useful information among relevant relations and pushing away irrelevant relations in the vector space.

\subsection{Entity Updating} 
\label{sec:eu}

Given a KG, this component updates the initial embeddings of entities via preserving the structural information. 
Unlike conventional GCN-based KGC methods, we conduct entity aggregation operations in the relational space. 
The basic assumption is that the same entity may express various semantics with different relations.
For instance, one man in the family is a nurturing and caring father, but in the company, his role will be changed to a productive and constructive employee. 
Therefore, the impact of relations should be considered for entity updating.

To do so, we leverage a weight-free GCN with an attention mechanism. Next, we will introduce the weight-free aggregation and the attention calculation, followed by the interpretability discussion. Note that there are also no additional parameters to compute attention scores.

\subsubsection{Weight-free Aggregation}
Here, we first give a general aggregation function $f_{ent}(\cdot)$ to describe the updating process of the $l$-th layer GCNs. Then, we will instantiate the exact function of aggregation with attention. Formally, we have:
\begin{equation}
\begin{split}
\mathbf{h}^{l+1}_t
&=f_{ent}\left(
    \left\{(\mathbf{h}^l_s,\mathbf{z}^l_r,\mathbf{h}^l_t)|(s,r)\in \mathcal{N}_t
    \right\}
    \right)
\end{split}
\end{equation}
where $\mathbf{h}^l_s$, $\mathbf{z}^l_r$ and $\mathbf{h}^l_t$ denote the hidden states of source entity $s$, relation $r$ and tail entity $t$ after $l$-th iteration in GCN layers, respectively.
Here, $\mathbf{h}^0_s=\mathbf{e}_s, \mathbf{z}^0_r=\mathbf{e}_r, \mathbf{h}^0_t=\mathbf{e}_t$.
$\mathcal{N}_t$ denotes the immediate neighbor relation-entity pairs of the tail entity $t$.
$f_{ent}(\cdot)$ is the entity updating function that integrates neighborhood information into the tail entity $t$. 

To guide the above aggregation with relation-specific semantics, we project each entity into the relational space via a composition operation:
$\phi: \mathbb{R}^d \times \mathbb{R}^d \rightarrow \mathbb{R}^d$.
That is, we view entity and relation as an associated combination. Herein, we apply diagonalization constraints to the relation hidden states and use such sparse representation as the projection matrix. Formally, we define $\phi$ as Hadamard product, i.e., the element-wise product of relation-entity pair:

\begin{figure*}
 \centering
 \includegraphics[width=1.0\textwidth]{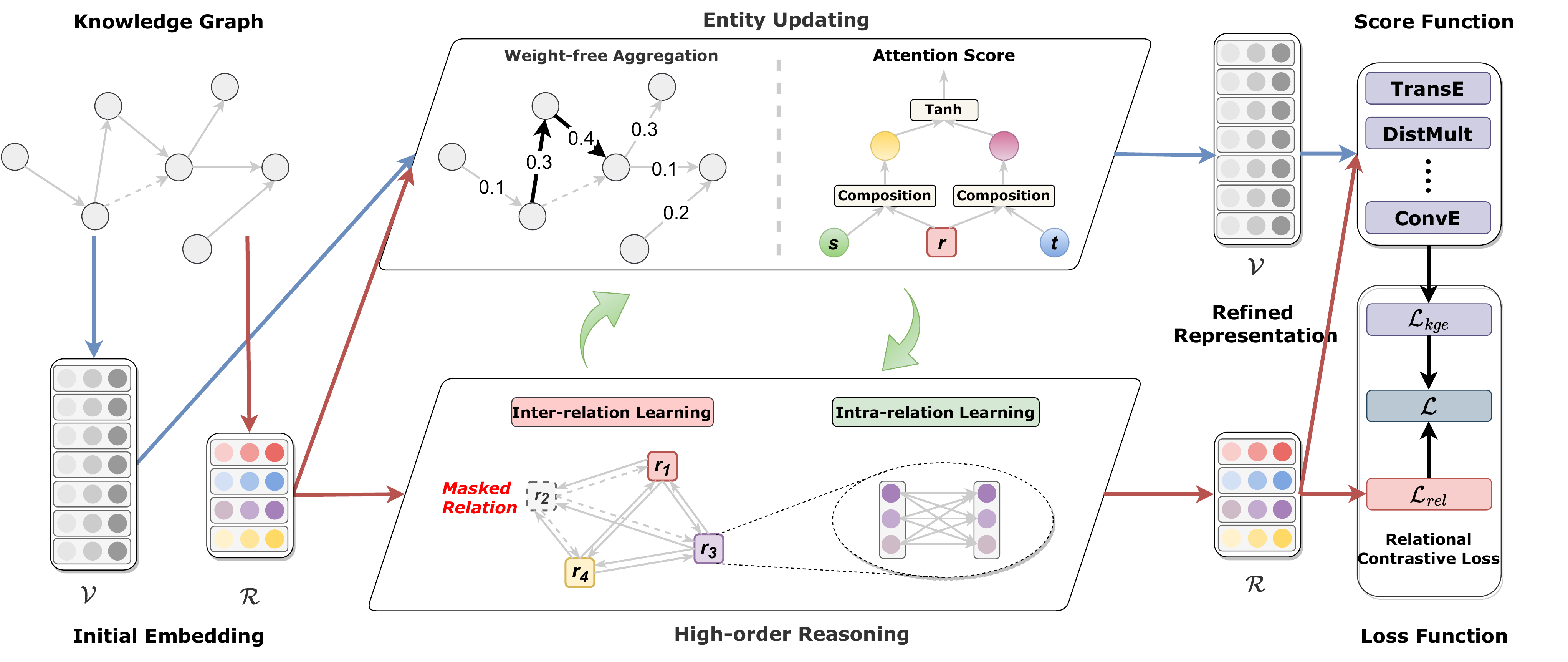}
 \caption{The overall framework of HoGRN. 
 The values on the edge of the entity updating part represent the possible attention scores, and the dotted lines in the inter-relation learning module represent the relations that may be randomly masked during the training stage. 
 HoGRN first iterates the entity updating and high-order reasoning process multiple times for representation learning. 
 Then the optimized entity and relation representations are fed into the scoring function to realize the plausibility evaluation of the facts, combined with auxiliary relational contrastive loss.
 Here we assume that there are seven entities ($N=7$), four relations ($M=4$), and both entity and relation have three dimensions ($d=3$).}
 \label{fig:framework}
%  \vspace{-10pt}
\end{figure*}

\begin{equation}
\begin{cases} 
\phi(\mathbf{h}^l_s,\mathbf{z}^l_r)  = \mathbf{h}^l_s \odot \mathbf{z}^l_r = diag(\mathbf{z}^l_r) \mathbf{h}^l_s  \\
\phi(\mathbf{h}^l_t,\mathbf{z}^l_r)  = \mathbf{h}^l_t \odot \mathbf{z}^l_r = diag(\mathbf{z}^l_r) \mathbf{h}^l_t
\end{cases}
\label{eq:ent_map_rel}
\end{equation}

The reason why we adopt such a lightweight strategy is on account of sparse scene representation learning, rather than as in previous work \cite{TransR,RGCN}, which typically introduces additional relation-specific matrices to complete the mapping of entities to the relational space. 
Of course, the general mapping function $\phi$ can be any other non-parameterized approaches, such as subtraction \cite{TransE} and circular-correlation \cite{HOLE}. We leave it to the future.

\subsubsection{Attention Calculation}
Furthermore, We introduce a relation-aware attention mechanism to define an exact aggregation function in the above weight-free GCNs. This not only makes the model interpretable~\cite{nathani2019learning,QA-GNN}, but also can enhance the reasoning ability by selecting informative neighbors and filtering out noise~\cite{GraIL}.
Formally, we instantiate the entity updating process $f_{ent}(\cdot)$ as follows:
\begin{equation}
\mathbf{h}^{l+1}_t
=\sum_{(s,r)\in \mathcal{N}_t}\frac{\alpha _{srt}}{\sqrt{d_sd_t}}\phi (\mathbf{h}^l_s,\mathbf{z}^l_r) 
\label{eq:ent_upd}
\end{equation}
where $d_s$ and $d_t$ are the degrees of entity $s$ and $t$ for scaling.
$\alpha _{srt}$ is the attention score used to measure the importance of a neighborhood relation-entity pair $(s,r)$ to the entity $t$, which is calculated by the following formula:
\begin{equation}
\alpha _{srt}=tanh \left [\phi ^T (\mathbf{h}^l_s,\mathbf{z}^l_r)\phi (\mathbf{h}^l_t,\mathbf{z}^l_r) \right ]
\label{eq:att_scr}
\end{equation}
Note that we use $tanh(\cdot)$ instead of the conventional $softmax(\cdot)$ here, mainly because the hyperbolic tangent function can help adaptively aggregate low and high-frequency signals from neighborhoods~\cite{FAGCN}.
To the best of our knowledge, this is the first GCN-based KGC model that maps both head and tail entities to the relational space for attention score calculation.

\subsubsection{Interpretability}
This strategy can help to model a holistic relation path between the head and tail entity, on which the attention score can be used to discover essential paths and filter out irrelevant parts.
We take the path with a high attention score as the reasoning process for the missing fact prediction.
Here, we assume that there exists the path set $\mathcal{P}_{s \rightarrow t}$ between the entities $s$ and $t$, which is composed of a series of paths $p := s \rightarrow r^1 \rightarrow r^2 \cdots \rightarrow r^l \rightarrow t$ ( $l$ is the path length and $r^1, r^2, \cdots, r^l$ are relations on the path).
According to Equation~\ref{eq:ent_upd} and Equation~\ref{eq:ent_map_rel}, we can obtain the following information flow:
\begin{equation}
\mathbf{h}^s_t
=\sum_{p \in \mathcal{P}_{s \rightarrow t}} \alpha _{r^1} \mathbf{z}_{r^1} \odot \alpha _{r^2} \mathbf{z}_{r^2} \odot \cdots \odot \alpha _{r^l} \mathbf{z}_{r^l} \odot \mathbf{h}_s
\label{eq:rel_path}
\end{equation}
where $\mathbf{h}^s_t$ is the information that entity $t$ can absorb from entity $s$. $\alpha _{r^1}, \alpha _{r^2}, \cdots, \alpha _{r^l}$ are normalized attentions on relation paths, which can help screen valuable and interpretable path.

Clearly, we have not introduced any additional parameters in Equation \ref{eq:ent_upd} and \ref{eq:att_scr}, except for the embeddings of entities and relations, which are necessary and the goal of KGC models.
There are some works \cite{SGC,LightGCN} that have demonstrated the advantages of simplifying GCNs.
This lightweight entity updating paradigm further improves computational efficiency in sparse KGC while providing interpretable relation paths for predicted results.

\subsection{High-order Reasoning} 
\label{sec:rr}

Given relation embeddings only, this component aims at improving them by capturing both endogenous correlation among relations (i.e., inter-relation learning) and self-enhancement (i.e., intra-relation learning).
Most conventional GCN-based KGC models are limited to entity updating \cite{DGCN,RGCN,WGCN,GraIL} with less exploration of relations.
However, relations play an essential role in rule-based reasoning.
Taking a simple rule as an example, $\forall  x,y,z \in \mathcal{V}$, if there exist facts $(x, hasMother, y)$ and $(y$, $hasHusband, z)$, then we have $(x, hasFather, z)$.
In other words, new facts are more probable inferred from those high-quality relation embeddings that preserve the endogenous correlation among them.
Such entity-agnostic information can help to extract generalized inference knowledge and mitigate the sparsity issue of the KGC task.

The above two sub-module design is inspired by the latest progress in the field of computer vision \cite{Mixer}.
Specifically, we first use the \textit{\textbf{inter-relation learning}} module to update the relations with the same dimensions. 
Second, we introduce \textit{\textbf{intra-relation learning}} to achieve self-renewal among relation's different dimensions. 
By iterating the above operations, the whole endogenous correlation learning process is completed.

\begin{figure}[t]
 \centering
 \includegraphics[width=0.89\columnwidth]{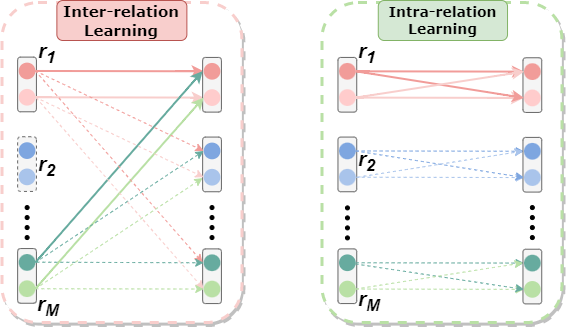}
%   \vspace{-10pt}
    \caption{
    The information flow procedure in the high-order reasoning component of HoGRN.
    Here, we show two dimensions of each relation and highlight $r_1$ with a bold solid line to make the update process more intuitive. 
    Illusory $r_2$ implies the relation that may be masked during the training stage.
    }
%  \vspace{-10pt}
 \label{fig:flow}
\end{figure}

For clarity, we first give a general update paradigm for relations involving the overall formula (Equation~\ref{sec:rg1} and \ref{sec:rg4}) and two compositional functions for each sub-module (Equation~\ref{sec:rg2} and \ref{sec:rg3}). 
As an intuitive demonstration, a visual illustration of this part is presented in Figure \ref{fig:flow}.
Then, we will instantiate these functions. 
Finally, we describe two strategies for better optimization. 
Before the introduction of the two sub-module, here is the general overall update paradigm:
\begin{equation}
\mathbf{z}^{l+1}_r = f_{rel} \left (\left \{\mathbf{z}^l_k|k \in \mathcal{R} \right \} \right )
\label{sec:rg1}
\end{equation}
where $f_{rel}(\cdot)$ is the high-order reasoning function composed of the following two sub-modules in this paper.

\subsubsection{Inter-relation Learning} 
Formally, the inter-relation learning process is defined as follows. Note that the updating is confined within the same dimension between different relation embeddings.
\begin{equation}
\begin{split}
z'_r(i)
& = f^1_{rel}\left (z^l_k(i) \right ), 1 \leqslant i \leqslant d, k \in \mathcal\mathcal{R}
\end{split}
\label{sec:rg2}
\end{equation}
where $z'_r(i)$ is $i$-th dimension of relation hidden states $\mathbf{z}'_r$, the updated representations of relation $r$ by inter-relation learning module. We remove the layer-wise upper-script for clarity.

\subsubsection{Intra-relation Learning}
The intra-relation learning module completes the following calculation:
\begin{equation}
z^{l+1}_r(i) = f^2_{rel}\left (z'_r(j) \right ), 1 \leqslant i,j \leqslant d
\label{sec:rg3}
\end{equation}
Finally, the relation representation is updated as follows:
\begin{equation}
z^{l+1}_r(i) = f^2_{rel} \left (f^1_{rel}\left (z^l_k(j) \right ) \right), 1 \leqslant i,j \leqslant d, k \in \mathcal\mathcal{R}
\label{sec:rg4}
\end{equation}

\subsubsection{Instantiation of High-order Reasoning Function}
With this progressive update strategy, each element of relations can fully absorb information from different dimensions of other relations.
We define the specific calculation method of the inter-relation learning module $f^1_{rel}(\cdot)$ as follows:
\begin{equation}
\mathbf{z}'_r = \left (\sigma (
[\mathbf{z}^{l}_1, \mathbf{z}^{l}_2, ..., \mathbf{z}^{l}_M]
\mathbf{W}^l_{1})\mathbf{W}^l_{2} \right )(*,r)
\end{equation}
where $\mathbf{W}^l_{1} \in \mathbb{R}^{M \times F_1}$,  $\mathbf{W}^l_{2} \in \mathbb{R}^{F_1 \times M}$ are weight matrices.
$\sigma(\cdot)$ is the activation function.
$\mathbf{X}(*,r)$ means the $r$-th column vector of the matrix $\mathbf{X}$.

For intra-relation learning module $f^2_{rel}(\cdot)$, we define it as follows:
\begin{equation}
\mathbf{z}^{l+1}_r= \mathbf{W}^{lT}_{4}\sigma  (\mathbf{W}^{lT}_{3}\mathbf{z}'_r )
\end{equation}
where $\mathbf{W}^l_{3} \in \mathbb{R}^{d \times F_2}$ and $\mathbf{W}^l_{4} \in \mathbb{R}^{F_2 \times d}$ are weight matrices.

\begin{figure}[t]
 \centering
 \includegraphics[width=0.69\columnwidth]{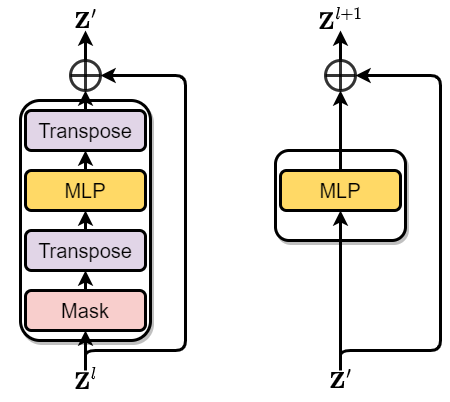}
%   \vspace{-10pt}
    \caption{
    The architecture of the high-order reasoning component, where the left box is the inter-relation learning module, and the right box is the intra-relation learning module.
    }
%  \vspace{-10pt}
 \label{fig:rel}
\end{figure}

The above strategies can be conveniently implemented in matrix form for efficient calculation as follows:
\begin{equation}
\begin{cases} 
\mathbf{Z}' = \mathbf{Z}^l + \left (\sigma((\mathbf{Z}^l)^T\mathbf{W}^l_{1})\mathbf{W}^l_{2} \right )^T \\
\mathbf{Z}^{l+1} = \mathbf{Z}' +  \sigma \left (\mathbf{Z}'\mathbf{W}^l_{3} \right )\mathbf{W}^l_{4}
\end{cases}
\end{equation}
where $\mathbf{Z}'$ is the intermediate outputs of inter-relation learning, and $\mathbf{Z}^{l+1} \in \mathbb{R}^{M \times d}$ are the final matrix representation of relations enhanced by intra-relation learning at the $l+1$ layer.
Here, we add the skip-connection to facilitate the possible deep stacking of this component \cite{ResNet,Mixer}.
We present a schematic of the calculation flow in Figure \ref{fig:rel}.

Instead of weight-free learning in the entity updating part, we introduce necessary parameters into high-order reasoning.
The primary consideration is that the density of information connected to relations is much richer than that of entities with only a tiny amount of neighbor information available in a sparse KG. 
Therefore, adding appropriate parameters is beneficial for learning high-quality relation representations. 
In addition, the number of relations in most KGs is limited, and its magnitude is significantly lower than that of entities, so the parameters added here are minimal and will not burden the model training.

\subsubsection{Optimization Strategy}
Clearly, according to Equation~\ref{sec:rg4}, each embedding dimension of a relation can be updated by any other embedding dimension. This may raise the issue of over-smoothing. 
That is, too much information fusion makes all relation vectors similar. 
To facilitate the optimization focusing on more informative and distinct features, we have designed the following two strategies: Stochastic Relation Masking and Relational Contrastive Constraints.

\textit{Stochastic Relational Masking} is inspired by the training strategy of masking input tokens in natural language processing \cite{Tranformer,BERT}, which can force the information fusion in self-attention focus on informative tokens via reconstruction. Here, we design a stochastic relational masking mechanism, defined as $\texttt {mask}(\cdot)$:
\begin{equation}
\mathbf{\hat{Z}}=\texttt {mask}(\mathbf{Z}^{l})
\end{equation}
We then replace $\mathbf{Z}^{l}$ with $\mathbf{\hat{Z}}$ as the inputs of the high-order reasoning component.
By randomly discarding a certain proportion of relations during the training stage, the model is forced to learn the endogenous correlation better.
Specifically, this strategy has twofold advantages.
On the one hand, this strategy forces the reserved relations to absorb information from more relevant ones in the remaining part; 
on the other hand, the randomly masked relations will need to rely more on the remnant with solid correlation to assist its own reconstruction.

\textit{Relational Contrastive Constraints.} In order to avoid possible trivial results where the distance between relation embeddings is too close in the vector space, we use mutual information to regularize their representations, which can be conveniently realized through contrastive learning \cite{CL}.
Specifically, we use InfoNCE \cite{InfoNCE} loss as a constraint on the learned relation representation as follows:

\begin{equation}
\mathcal{L}_{rel}
=\sum_{r \in \mathcal{R}} - \log \frac{exp(s(\mathbf{z}^{L}_r,\mathbf{z}^{L}_r)/ \tau)}{\sum_{r' \in \mathcal{R}}exp(s(\mathbf{z}^{L}_r,\mathbf{z}^{L}_{r'})/ \tau)}
\end{equation}
where $s(\cdot)$ measures the similarity between the two relations' representation, which is set as a cosine similarity function; $\tau$ is a temperature coefficient, the hyper-parameter in the $softmax$ function.
Note that we use the output of the last layer ($l=L$) of the HoGRN's high-order reasoning component to calculate this loss. 

\subsection{Score Function}
\label{sec:sf}
We successively perform entity updating and high-order reasoning parts and iterate this procedure for $L$ times to complete the information propagation in the sparse KG. 
Then, the plausibility evaluation of the fact $(s,r,t)$ is completed by inputting the refined representation $\mathbf{h}^L_s$, $\mathbf{z}^L_r$, and $\mathbf{h}^L_t$ into the scoring function.
HoGRN is compatible with almost arbitrary embedding-based KGC models' score functions. Here, we choose the following state-of-the-art models:
\begin{itemize}
    \item  TransE \cite{TransE}:
    \begin{equation}
        f(\mathbf{h}^L_s,\mathbf{z}^L_r,\mathbf{h}^L_t) = -||\mathbf{h}^L_s + \mathbf{z}^L_r - \mathbf{h}^L_t||_{l_{1/2}}
    \end{equation}
    where the subscript $l_{1/2}$ denotes either $L_1$-norm or squared $L_2$-norm. \footnote{In the experiment we use $L_1$-norm because of its better performance.}
    \item DistMult \cite{DistMult}:
    \begin{equation}
        f(\mathbf{h}^L_s,\mathbf{z}^L_r,\mathbf{h}^L_t) = {\mathbf{h}^{L}_s}^T diag(\mathbf{z}^L_r) \mathbf{h}^L_t
    \end{equation}
    where $diag(\cdot)$ is the operation that transforms a vector into a diagonal matrix.
    \item ConvE \cite{ConvE}:
    \begin{equation}
    \begin{split}
        f(\mathbf{h}^L_s,\mathbf{z}^L_r,\mathbf{h}^L_t) =  {\mathbf{h}^{L}_t}^T ReLU(\mathbf{W} \\ \bullet \ vec(ReLU(concat(\bar{\mathbf{h}}^L_s,\bar{\mathbf{z}}^L_r)\star \mathbf{\Omega})))
    \end{split}
    \end{equation}
    where $\bar{\mathbf{h}}^L_s$, $\bar{\mathbf{z}}^L_r$ denote a 2D reshaping of $\mathbf{h}^L_s$ and $\mathbf{z}^L_r$. 
    $concat(\cdot)$ is the concatenation operation that splices vectors together.
    $vec(\cdot)$ is the vectorization operation used to convert a matrix to a vector.
    $\star$ and $\mathbf{\Omega}$ denote the convolution operator and a set of filters, respectively.
\end{itemize}

\subsection{Training Regime} 
\label{sec:tr}

Following the previous work\cite{RGCN,CompGCN}, we first define a principle loss as a binary cross entropy (BCE) loss based on the above scoring function as follows:
\begin{equation}
\begin{split}
\mathcal{L}_{kge}=-\frac{1}{N|\mathcal{T}'|}\sum_{(s,r,t) \in \mathcal{T}', o \in \mathcal{V}} y \log l(f(\mathbf{h}^L_s,\mathbf{z}^L_r,\mathbf{h}^L_o)) \\ 
+ \ (1-y) \log (1-l(f(\mathbf{h}^L_s,\mathbf{z}^L_r,\mathbf{h}^L_o)))
\end{split} 
\end{equation}
where $l(\cdot)$ is the logistic sigmoid function, $o$ is all of the possible tail entity in the entity set $\mathcal{V}$, and $y$ is an indicator with $y=1$ iff $(s,r,o) \in \mathcal{T}'$.

Then, we take the relational contrastive loss as an auxiliary part and combine it with the former to obtain the final optimization function.
Finally, we use the following loss function for KGC models' end-to-end training and optimization:
\begin{equation}
\mathcal{L}=\mathcal{L}_{kge}+\lambda \mathcal{L}_{rel}
\end{equation}
where $\lambda$ is the hyper-parameter to balance the influence of auxiliary loss.

\subsection{Disscusion}
Here, we give a further discussion compared with other GCN-based KGC models. Typically, these GCN-based KGC models adopt the encoder-decoder architecture. Firstly, they use GCNs as the encoder to encode graph structure information to optimize entity and relation representations. Then, they adopt a generic embedding-based KGC model (e.g., TransE \cite{TransE}, DistMult \cite{DistMult}, ConvE \cite{ConvE}) as the decoder to reconstruct graph structures \cite{RGCN}. 
Thus, these approaches can strengthen the embedding-based KGC model with better performance by modeling the graph structures.

HoGRN not only simplifies the aggregation process due to the sparse connections among entities but also captures high-order reasoning for relations considering their rich information. As shown in Table~\ref{tab:kgcn}, we present several representative GCN-based KGC models: RGCN \cite{RGCN}, WGCN \cite{WGCN}, and CompGCN \cite{CompGCN}. HoGRN takes more consideration of relations when updating entities compared to the previous work, and this improvement helps to find valuable relation paths to make the model interpretable (see Formula \ref{eq:rel_path} for further clarification). In particular, when updating relations, HoGRN explicitly learns the relations' endogenous correlation to strengthen model inference, which is especially important in sparse KG scenarios.

\begin{table*}
\centering
\caption{
Statistics of the datasets, where degrees indicate the sparsity difference among datasets.
}
\scalebox{0.94}{
\begin{tabular}{l r r r r r r c c}
\toprule[1pt]
\multirow{2}{*}{\textbf{Dataset}} & \multirow{2}{*}{\textbf{\# Entities}} & \multirow{2}{*}{\textbf{\# Relations}} & \multicolumn{4}{c}{\textbf{Triplets}} &  \multicolumn{2}{c}{\textbf{Degrees}} \\ 
\cmidrule(r){4-7} \cmidrule(r){8-9}
&    &    & \textbf{\# Train}  & \textbf{\# Valid} & \textbf{\# Test} & \textbf{Total} & \textbf{average} & \textbf{median}  \\
\midrule
\textbf{NELL23K}        & 22,925 & 200 & 25,445 & 4,961 & 4,952 & 35,358 & 2.21 & 1  \\ 
\textbf{WN18RR} & 40,943 & 11 & 86,835 & 3,034 & 3,134 & 93,003 & 2.29 & 2  \\
\textbf{WD-singer}      & 10,282 & 135 & 16,142 & 2,163 & 2,203 & 20,508 & 2.35 & 2  \\ 
\textbf{FB15K-237-10\%} & 11,512 & 237 & 27,211 & 15,624 & 18,150 & 60,985 & 5.84 & 4  \\
\textbf{FB15K-237-20\%} & 13,166 & 237 & 54,423 & 16,963 & 19,776 & 91,162 & 7.53 & 5  \\
\textbf{FB15K-237-50\%} & 14,149 & 237 & 136,057 & 17,449 & 20,324 & 173,830 & 13.04 & 9  \\
\textbf{FB15K-237} & 14,541 & 237 & 272,115 & 17,535 & 20,466 & 310,116 & 22.32 & 16  \\
\bottomrule[1pt]
\end{tabular}}
% \vspace{+1pt}
\label{tab:dataset}
\end{table*}

\begin{table*}
\centering
\caption{
Performance comparison of link prediction task on three different datasets. All metrics are multiplied by 100.
In brackets is the relative improvement of HoGRN compared with CompGCN under the same scoring function.
Here, we use bold to highlight the highest value and underline the runner-up.}
\scalebox{0.97}{
\begin{tabular}{l ccc ccc ccc}
\toprule[1pt]
\multirow{2}{*}{\textbf{Method}} & 
\multicolumn{3}{c}{\textbf{NELL23K}} &
\multicolumn{3}{c}{\textbf{WD-singer}} & 
\multicolumn{3}{c}{\textbf{FB15K-237-10\%}} \\
\cmidrule(r){2-4}
\cmidrule(r){5-7}
\cmidrule(r){8-10}
& \textit{MRR}  & \textit{Hits@3} & \textit{Hits@10}  
& \textit{MRR}  & \textit{Hits@3} & \textit{Hits@10}
& \textit{MRR}  & \textit{Hits@3} & \textit{Hits@10} \\
\midrule
\textbf{RotatE}   
&  18.33   &  19.66   &  30.06  
&  33.13   &  36.27   &  42.90  
&  13.60   &  14.38   &  24.46 \\
\textbf{TuckER}   
&  20.77   &  22.75   &  36.13  
&  37.05   &  40.45   &  45.73 
&  16.34   &  17.20   &  27.54  \\
\midrule
\textbf{TransE}   
&  16.03   &  17.98   &  30.15  
&  32.87   &  39.79   &  49.37  
&  14.85   &  15.69  &  25.30  \\
\textbf{DistMult}   
&  17.83   &  19.68   &  29.90  
&  31.56   &  34.14   &  38.31 
&  14.36   &  14.99   &  24.11  \\
\textbf{ConvE}   
&  22.43   &  24.31   &  37.90  
&  36.41   &  39.58   &  45.98  
&  16.53   &  17.41   &  27.74  \\
\midrule
\textbf{CompGCN-TransE}   
&  17.14   &  18.41   &  30.10  
&  34.55   &  39.92   &  48.77  
&  15.63   &  16.41   &  26.83  \\
\textbf{CompGCN-DistMult}   
&  19.05   &  21.57   &  33.69  
&  35.24   &  39.36   &  46.98  
&  13.93   &  14.56   &  23.67  \\
\textbf{CompGCN-ConvE}   
&  \underline{24.16}   &  \textbf{26.80}   &  \textbf{40.24}  
&  37.04   &  41.49   &  48.46  
&  16.61   &  17.43   &  28.28  \\
\midrule
\textbf{HoGRN-TransE}   
&  20.52   &  22.06   &  34.91 
&  \underline{37.85}   &  \textbf{44.40}   &  \textbf{50.95}  
&  \textbf{17.15}   &  \textbf{18.07}   &   \underline{28.96}  \\
\textit{--Rela. Impr.}   
&  \textit{(+19.7\%)}    &  \textit{(+19.8\%)}    &  \textit{(+16.0 \%)}
&  \textit{(+\ 9.6\%)}     &  \textit{(+11.2\%)}     &  \textit{(+\ 4.5\%)} 
&  \textit{(+\ 9.7\%)}     &  \textit{(+10.1\%)}     &  \textit{(+\ 7.9\%)}  \\
\textbf{HoGRN-DistMult}   
&  20.81   &  23.01   &  35.86  
&  37.50   &  41.04   &  47.39  
&  16.14   &  17.05   &  26.69  \\
\textit{--Rela. Impr.}   
&  \textit{(+\ 9.2\%)}    &  \textit{(+\ 6.7\%)}     &  \textit{(+\ 6.4\%)}
&  \textit{(+\ 6.4\%)}     &  \textit{(+\ 4.3\%)}     &  \textit{(+\ 0.9\%)} 
&  \textit{(+15.9\%)}     &  \textit{(+17.1\%)}     &  \textit{(+12.8\%)}  \\
\textbf{HoGRN-ConvE}   
&  \textbf{24.56}   &  \underline{26.68}   &  \underline{39.98}  
&  \textbf{39.07}   &  \underline{42.67}   &  \underline{48.80}  
&  \textbf{17.15}   &  \underline{17.95}   &  \textbf{29.22}  \\
\textit{--Rela. Impr.}   
&  \textit{(+\ 1.7\%)}    &  \textit{(-\ 0.4\%)}    &  \textit{(-\ 0.6\%)}
&  \textit{(+\ 5.5\%)}     &  \textit{(+\ 2.8\%)}     &  \textit{(+\ 0.7\%)} 
&  \textit{(+\ 3.3\%)}     &  \textit{(+\ 3.0\%)}     &  \textit{(+\ 3.3\%)}  \\
\bottomrule[1pt]
\end{tabular}
}
% \vspace{-11pt}
\label{result}
\end{table*}

\section{Experiments}
\label{sec:experiments}

% This is an experiment.

In this section, we conduct extensive experiments on multiple datasets to demonstrate the rationality and validity of our proposed model HoGRN. 
Specifically, we want to design a series of experiments that answer the following questions:
\begin{description}
\item[RQ1] How does HoGRN perform on sparse KGs compared to current advanced KGC methods?
\item[RQ2] Do different components of HoGRN affect the performance positively?
\item[RQ3] How does HoGRN perform at different levels of KG sparsity?
\item[RQ4] What is the influence of optimization hyper-parameters, e.g., relational masking ratio, embedding dimensions?
\item[RQ5] How do HoGRN's interpretable results look like?
\end{description}

\subsection{Experimental Settings}

\subsubsection{\textbf{Datasets}}
This paper takes three sparse datasets constructed in DacKGR \cite{DacKGR} to evaluate our model.
Among them, NELL23K and WD-singer are two datasets sampled from NELL \cite{NELL} and Wikidata \cite{Wikidata}.
FB15K-237-10\% is a subset of the widely used KGC dataset FB15K-237 \cite{FB15K237}, where 10\% denotes that 90\% triplets are removed from the original dataset.
These three datasets have high sparsity and rich relations to support our experimental verification.
The detailed statistics are shown in Table \ref{tab:dataset}. We present the average and median out-degree of entities to give a direct impression of KG sparsity. A high degree denotes a low sparsity. Besides, we also show the statistics of WN18RR~\cite{ConvE}, FB15K-237, and its various subsets, which are used in further analysis on RQ3.

\subsubsection{\textbf{Baseline Models}}
{For fair comparison}, we mainly compare HoGRN with the GCN-based KGC method CompGCN because it achieves state-of-the-art performances and is similar to our design. Also, we compare different types of embedding-based KGC models, including translation-based (TransE, RotatE), semantic matching-based (DistMult, TuckER), and neural network-based (ConvE).
The details of these baseline models are as follows:
\begin{itemize}[leftmargin=*]
\item \textbf{TransE} \cite{TransE}:
This is a representative translation model which assumes that the sum of the embeddings of the head entity and the relation should be close to the embedding of the tail entity.
\item \textbf{DistMult} \cite{DistMult}:
It proposes a simple bilinear formulation to incorporate the embedding of entities and relations and measure the plausibility of facts through semantic matching.
\item \textbf{ConvE} \cite{ConvE}: 
This method uses 2D convolution over the embedding of entities and relations and multiple layers of nonlinear features to express semantic information.
\item \textbf{RotatE} \cite{RotatE}: 
This is also a translation-based model that views relations as the rotation from source entities to target entities in the complex vector space.
\item \textbf{TuckER} \cite{TuckER}: 
This strategy is a semantic matching-based model, which introduces three-way Tucker tensor decomposition to learn the embedding of entities and relations.
\item \textbf{CompGCN} \cite{CompGCN}:
The model uses entity-relation combination operation to integrate relation embedding into GCNs' formulation and then uses the embedding-based model as the scoring function, which achieves excellent performance.
\end{itemize}

Note that the multi-hop reasoning-based approaches such as MultiHopKG \cite{MultiHopKG} and DacKGR \cite{DacKGR} are not chosen here, because they have demonstrated inferior performance compared to the embedding-based approaches \cite{lin2018multi,fu2019collaborative,MultiHopKG,DacKGR} and are difficult to train stably on sparse settings.

\subsubsection{\textbf{Evaluation Metrics}}
Following previous studies \cite{CompGCN,DacKGR}, we evaluate all models on the link prediction task to reflect their performance in sparse KGC scenarios. That is,  to predict possible tail entities given head entities and relations.
Similarly, we use mean reciprocal rank (MRR) and the correct tail entities ranking proportion in top K (Hits@K) as evaluation metrics with a filtered setting \cite{TransE,RGCN}.

\subsubsection{\textbf{Implementation Details}}
We follow the hyper-parameter settings as CompGCN \cite{CompGCN}, and perform a grid search to acquire the best hyper-parameter configuration.
Specifically, the learning rate is tuned in $\{1e{-3}, 1e{-4}\}$, and the batch size is tuned among $\{128, 256\}$.
In addition, the number of GCN layer $L$ is searched in $\{1,2,3\}$ and the entity and relation's dimension $d$ is selected from $\{100,150,200\}$. 
In addition, we choose stochastic relational masking ratio from $\{0.0, 0.1, 0.2, 0.3\}$ and tune the temperature coefficient $\tau$ within the ranges of $\{0.2, 0.5, 1.0, 2.0\}$.
The trainable parameters of our model, including entity and relation embedding, are initialized with the Xavier \cite{Xavier} and optimized by Adam \cite{Adam}.
We use GELU \cite{GELU,Mixer} as the activation function in the high-order reasoning part.
Both CompGCN and HoGRN use the Hadamard product as the composition operation; other choices we reserve for future work.
Along with the setup of CompGCN, TransE, DistMult, and ConvE are used as scoring functions for these two models, respectively.
In all cases, we apply an early stop strategy on the validation set with the patience of 25 epochs and report the optimal state based on the metric MRR.

\subsection{\textbf{Overall Comparison (RQ1)}}
The overall performance of HoGRN, as well as baseline models, are presented in Table \ref{result}. We can see that:
\textbf{(1)} our proposed HoGRN achieves the best performance across all sparse datasets, evaluation metrics, and even different backbone scoring functions, except Hits@3 and Hits@10 on NELL23K (the scores are very comparable). 
Quantitatively, the improvements with respect to three metrics, MRR, Hits@3, and Hits@10, are 9.0\%, 8.3\%, and 5.8\% on average, respectively.
This validates the superiority of our proposed approach to capture high-order correlation among relations when updating entities and relations on sparse KGs.
\textbf{(2)} Almost all KGC methods perform unsatisfactorily in such sparse settings. This highlights the importance of the investigation of sparse KGC in a real-world scenario.
\textbf{(3)} Among all the embedding-based baseline models, ConvE achieves the most competitive performance in most cases. This may be attributed to ConvE's better representation learning ability by introducing appropriate parameters other than entity and relation embedding.
\textbf{(4)} Furthermore, when combined with GCNs (i.e., CompGCN), the performance of all scoring functions is almost improved. This illustrates the importance of preserving the graph structure information into entity and relation representations for better performance.

Interestingly, we notice that more advanced models may not achieve a better performance in sparse scenarios --- the classic TransE performs very competitively, especially on WD-singer, and the scores of CompGCN-DistMult on FB15K-237-10\% have degraded compared with the vanilla DistMult.
The possible reason is that sophisticated architecture may increase the training difficulty and thus perform not well. 
By contrast, our designs simplify the entity updating strategy and pay more attention to the high-order reasoning, which can stably improve the backbone KG embeddings --- with HoGRN, even the most basic embedding-based model TransE achieves the best performance on WD-singer and FB15K-237-10\%.

\begin{table}[t]
\centering
\caption{
Performance comparison of link prediction task on WN18RR. All metrics are multiplied by 100. * indicate that the results are taken from their raw paper ('-' denotes a missing value). Here, both CompGCN and HoGRN use ConvE as the scoring function.
}
% \vspace{+4pt}
\resizebox{0.8\columnwidth}{!}{
\begin{tabular}{l c c c}
\hline
\multicolumn{1}{l}{\textbf{Method}} 
& \textit{MRR}  & \textit{Hits@3} & \textit{Hits@10} \\
\hline

\textbf{TransE}\cite{TransE}* &  22.6   &  -   &  50.1  \\
\textbf{DistMult}\cite{DistMult}* &  43.0   &  44.0   &  49.0 \\
\textbf{ComplEx}\cite{ComplEx}* &  44.0   &  46.0  &  51.0 \\
\textbf{ConvKB}*\cite{ConvKB} &  24.9   &  41.7 &  52.4  \\
\textbf{SACN}\cite{WGCN}* &  47.0   &  48.0  &  54.0 \\
\textbf{HypER}*\cite{HypER} &  46.5   &  47.7  &  52.2 \\
\textbf{RotatE}\cite{RotatE}* &  47.6   &  49.2  &  \textbf{57.1}  \\ 
\textbf{TuckER}\cite{TuckER}* &  47.0   &  48.2   &  52.6  \\
\textbf{ConvR}*\cite{ConvR} &  47.5   &  48.9  &  53.7 \\
\hline
\textbf{ConvE}\cite{ConvE}* &  43.0   &  44.0  &  52.0 \\
\textbf{CompGCN}\cite{CompGCN}* &  \underline{47.9}   &  \underline{49.4}   &  54.6  \\
\hline
\textbf{HoGRN} &  \textbf{48.01}   &  \textbf{49.54}   & \underline{55.86} \\
\hline
\end{tabular}}
% \vspace{+2pt}
\label{tab:wr}
\end{table}

\subsection{\textbf{Ablation Study (RQ2)}}

\begin{figure}[t]
 \centering
 \includegraphics[width=0.99\columnwidth]{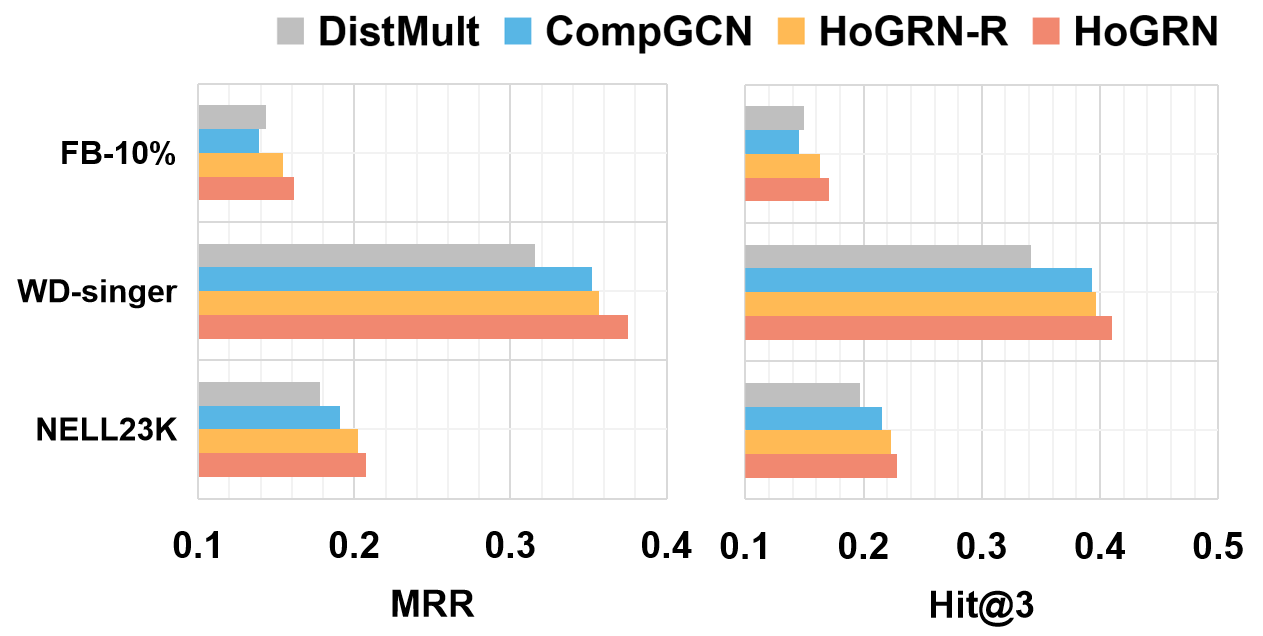}
%   \vspace{+2pt}
 \caption{Ablation study results in three sparse KGs.
 Here, both CompGCN and HoGRN take DistMult as the score function.
 }
%  \vspace{-10pt}
 \label{fig:ab}
\end{figure}

Our core idea to deal with the sparsity issue is the high-order reasoning component. In this section, we explore its impacts and mark the variant of removing the high-order reasoning component as \textbf{HoGRN-R}. For a fair comparison, we choose DistMult as the backbone scoring function and also involve CompGCN-DistMult (short for CompGCN here) as the baseline. Note that we didn't select ConvE as the backbone scoring function because the additional parameters make their optimizations a bit different.

The experimental results are shown in Figure \ref{fig:ab}.
From this figure, we can have the following observations:
\begin{itemize}[leftmargin=*]
\item After removing the high-order reasoning part, the performance of HoGRN-R on all three datasets declines. 
Specifically, compared with HoGRN, the MRR scores decrease by $\textit{2.82\%}$, $\textit{4.93\%}$, $\textit{4.28\%}$ in datasets NELL23K, WD-singer, and FB15K-237-10\%, respectively.
This indicates the effectiveness of the high-order reasoning component.
\item HoGRN-R still outperforms CompGCN with our weight-free GCNs to update entities only. 
The MRR of HoGRN-R exceeds CompGCN by $\textit{6.25\%}$, $\textit{1.16\%}$, $\textit{10.91\%}$ on three datasets, respectively. 
This demonstrates the positive impact of the advancement on our entity updating strategy.
\end{itemize}

\subsection{\textbf{Sparsity Analysis (RQ3)}}
\begin{figure}[t]
 \centering
 \includegraphics[width=0.99\columnwidth]{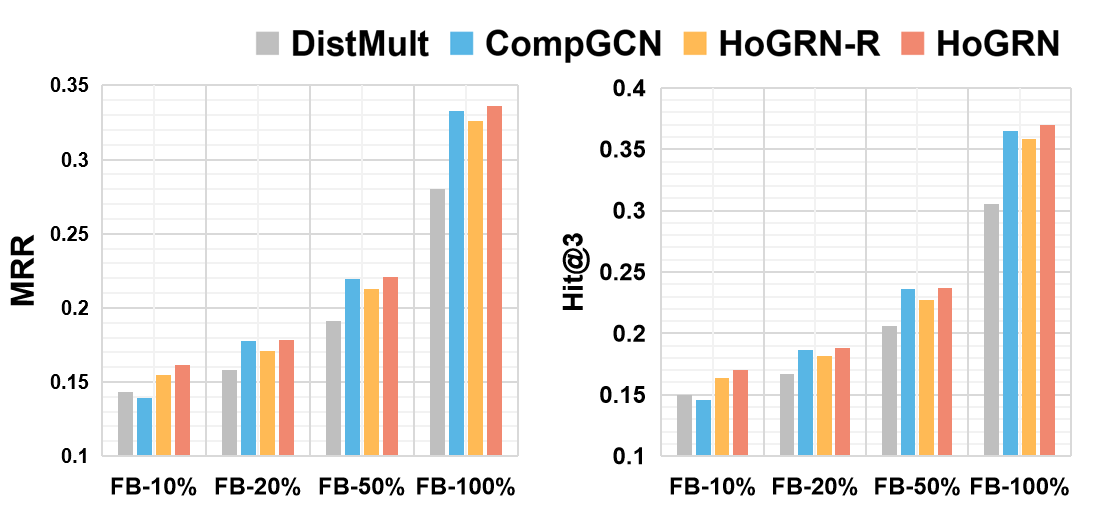}
%  \vspace{-10pt}
 \caption{Experimental results of a series of comparative models on FB15K-237 and its different sparsity subsets.
 }
%  \vspace{-10pt}
 \label{fig:sp}
\end{figure}

In this section, we analyze the sparsity impacts on the KGC task from two aspects. First, we compare more state-of-the-art models in a widely used dataset, WN18RR, which is also sparse (similar average and median degrees with NELL23K and WD-singer in Table~\ref{tab:dataset}), although the limited relations reduce the task difficulty. Second, we gradually remove a portion triples from FB15K-237 and explore the impacts of different graph sparsity levels. Note that 10\% denotes the removal of 90\% training triples, and 100\% denotes the original dataset.

\subsubsection{Comparison on Full WN18RR}

As shown in Table~\ref{tab:wr}, we can see that HoGRN still achieves the best or second-best performance on this full version of the widely used dataset across different evaluation metrics. Besides, although WN18RR is very sparse, the performance is better than that on those sparse datasets (i.e., NELL23K, WD-singer, FB15K-237-10\%) due to the fewer relations. That is, the number of relations becomes less, and their information is relatively affluent, which is helpful for inferring high-order rules for link prediction. This agrees with our component that composites high-order reasoning into GCN for sparse KGC. And that's why HoGRN also achieves the best performance in Full WN18RR.

\subsubsection{Comparison on Different Sparsity Levels}

To illustrate the impact of data sparsity on model performance, we further conduct experiments on FB15K-237 and its subsets of different sparsity. 
Specifically, we compare CompGCN and HoGRN with the scoring function DistMult.
The results are shown in Figure \ref{fig:sp}.
As can be seen from the figure:
\begin{itemize}[leftmargin=*]
    \item At the sparsest subset FB-10\%, HoGRN achieves significant improvements (the relative increase of MRR is $\textit{15.87\%}$ compared with CompGCN). 
    With the increase in data density, the improvements of HoGRN compared with CompGCN are relatively limited. It is noted that the performance of HoGRN-R (with weight-free GCN only) is better than CompGCN in the lowest sparsity but not ideal in other cases. We attribute the weight-free design may limit the performance of HoGRN on rich data, but it is convenient to add a learnable weight matrix if the KG is not very sparse.
    On the whole, this not only demonstrates the effectiveness of HoGRN in dealing with sparse KGC but also verifies that HoGRN will not harm the performance when there are sufficient training triples. 
    In fact, the improvements over the backbone KGC embeddings are still promising.
    \item Compared with the performance of HoGRN and its variant HoGRN-R, it can be seen that the high-order reasoning component can bring stable improvements (the relative gains of MRR are $\textit{4.46\%}$, $\textit{4.11\%}$, $\textit{3.61\%}$, and $\textit{3.04\%}$, respectively). 
    This shows that the design can adapt well to different sparsity situations and effectively improve the model's performance on the KGC task.
    \item Compared with DistMult, the GCN-based models can almost achieve better performance under different sparsity levels.
    This also shows the advantage of the GCN-based model in improving the quality of entity and relation representation. 
    In addition, the degradation performance of CompGCN on the most sparse subset of FB15K-237 also illustrates the necessity of simplifying the model architectures in the sparse scenario. We may explore various simplifying strategies in the future.
\end{itemize}

\subsection{\textbf{Hyper-parameter Study (RQ4)}}

\begin{figure*}[t]
 \centering
 \includegraphics[width=0.8\textwidth]{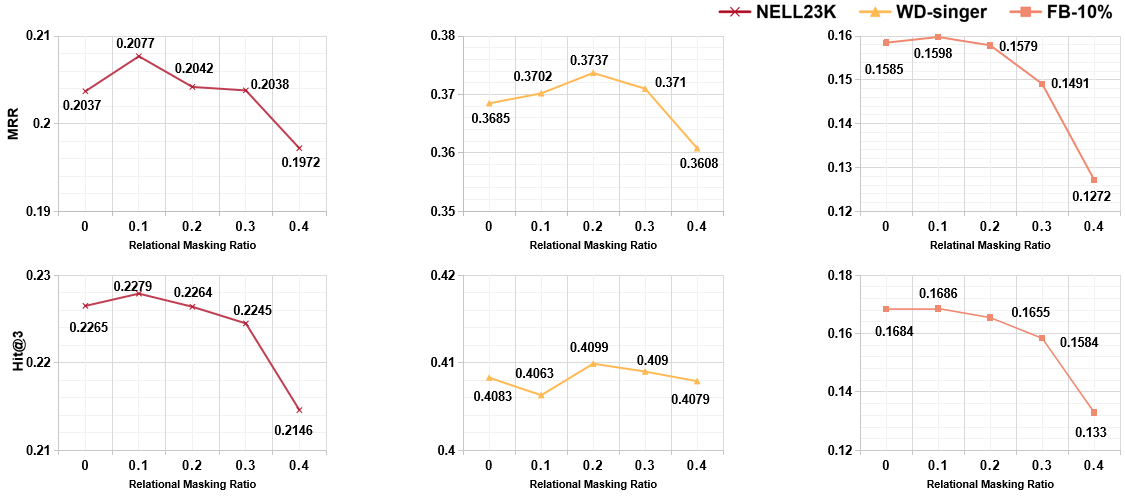}
%   \vspace{-5pt}
 \caption{Experimental results of HoGRN under different relational masking ratios. 
 }
%  \vspace{-15pt}
 \label{fig:mask1}
\end{figure*}

\begin{figure*}[t]
 \centering
 \includegraphics[width=0.8\textwidth]{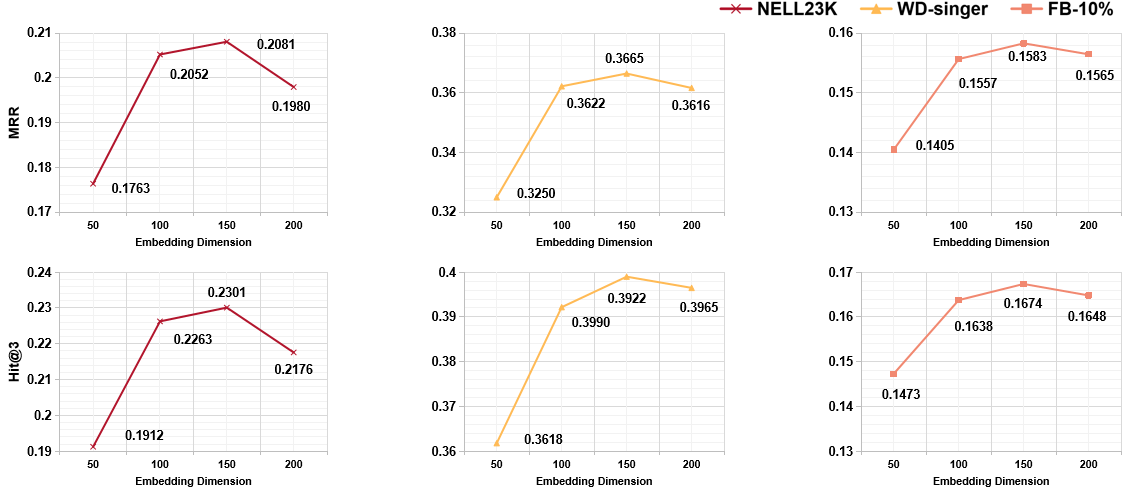}
%   \vspace{-5pt}
 \caption{Experimental results of HoGRN under different embedding dimensions.
 }
 \vspace{-5pt}
 \label{fig:dim1}
\end{figure*}

Here, we analyze the influence of two essential hyper-parameters on model performance: relational masking ratio and embedding dimension. Specifically, we fix other hyper-parameters to make a fair comparison. Similarly, DistMult is still used as HoGRN's scoring function for analysis.

\subsubsection{Relational Masking Ratio}

To further study the effect of the key design: stochastic relational masking in high-order reasoning component. We change the masking ratio between intervals [0,0.4] and report the model performance in Figure \ref{fig:mask1}.
Here, the batch size is set as 256, the number of GCNs' layers is fixed as two, and DistMult is selected as the scoring function.
From this figure, we can find that:
\begin{itemize}[leftmargin=*]
    \item Masking relations at a certain ratio during the training stage can effectively lead to performance improvements on all three datasets.
    Specifically, the performance on NELL23K, WD-singer, and FB15K-237-10\% are improved by $\textit{1.96\%}$, $\textit{1.41\%}$, and $\textit{0.82\%}$ on MRR, respectively (compared with the situation without the introduction of relational masking strategy, i.e., the relation masking ratio is set as 0).
    \item An appropriate dropout ratio of 0.1 or 0.2 usually yields good benefits on HoGRN. 
    This can be attributed to the fact that this strategy can force the model to better learn the relationship among relations so as to enhance the model's reasoning ability.
    \item When the relational masking ratio is greater than 0.3, the model performance will decrease significantly. 
    The possible reason is that when the relational masking ratio is too large, it is difficult to rely on the remaining relations to assist the reconstruction of missing ones, thus harming the performance of the model.
\end{itemize}

\begin{figure*}[t]
 \centering
 \includegraphics[width=0.89\textwidth]{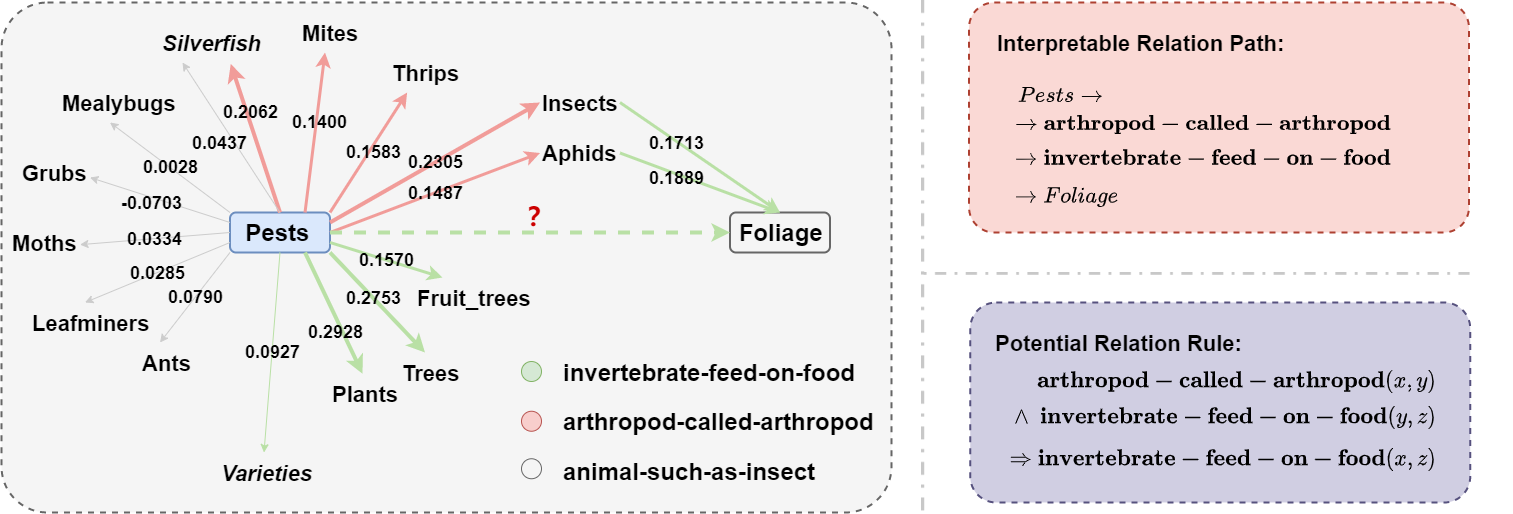}
%   \vspace{-10pt}
 \caption{A case from NELL23K. Here, lines of different colors correspond to different relations, and the thickness of lines reflects the difference in attention score. The dashed line corresponds to the fact to be predicted (from the test set), and solid lines correspond to known facts from the training set.
 }
%  \vspace{+5pt}
 \label{fig:cs}
\end{figure*}

\subsubsection{Embedding Dimension}

Here, we investigate the effect of embedding dimensions under the following configuration: the batch size is fixed to 256, the number of GCNs' layers equal to 2, and the relational masking ratio is set as 0. 
We test the performance of HoGRN in the embedding dimension set $\{50,100,150,200\}$.
The results are shown in figure \ref{fig:dim1}.
As can be seen from this figure, when DistMult is used as the scoring function with the former settings, generally, good performance can be achieved in the range [100, 200] of embedding dimensions, which is consistent with previous work \cite{CompGCN} on non-sparse KG. 
At the same time, when the embedding dimension is reduced to 50, the model's performance is significantly impaired. 
In other words, simply reducing the embedding dimension cannot improve the model's performance in sparse scenarios.
Therefore, this further highlights the critical value of the higher-order reasoning component and the simplification of GCN architecture in the entity updating component proposed by HoGRN.

\subsection{\textbf{Case Study (RQ5)}}

To give an intuitive impression of the rationality and validity of our model design, we take a fact in NELL23K as an illustration.
As shown in Figure \ref{fig:cs}, \textit{(Pests, invertebrate-feed-on-food, Foliage)} is an example of the test set where \textit{Pests} with high degree and \textit{Foliage} with low degree.
From the figure, we can acquire the following observations:
\begin{itemize}[leftmargin=*]
    \item Based on the attention scores, we can find valuable paths between the head entity and tail entity. For example, both two paths between \textit{Pests} and \textit{Foliage} linked by \textit{invertebrate-feed-on-food} and \textit{arthropod-called-arthropod} have high attention scores (other informative paths are omitted for clarity), which will contribute more to the prediction on the missing fact. Clearly, these paths are reasonable as a rule: given \textit{(x, arthropod-called-arthropod, y)} and \textit{(y, invertebrate-feed-on-food, z)}, we may have \textit{(x, invertebrate-feed-on-food, z)}. This not only shows that HoGRN can provide interpretability for predicted results but also demonstrates that HoGRN can effectively learn the relationships among relations --- invertebrates, including arthropods, they are closely correlated.
    \item The facts under relations \textit{invertebrate-feed-on-food} and \textit{arthropod-called-arthropod} have similar high attention scores, while \textit{animal-such-as-insect} has much lower scores. This suggests that the first two relations can provide more informative and exact semantics to understand entity \textit{Pests}. 
    \item In specific, the attention score of neighbor entity \textit{Silverfish} (a species of tiny insect, not fish) under \textit{arthropod-called-arthropod} is significantly higher than those entities under relation \textit{animal-such-as-insect}, although they are all related to the same head entity \textit{Pests}. This is because \textit{Silverfish} is exactly considered as household pests, while others like \textit{Mealybugs} is very rare (e.g., little information can be found in Wikipedia), or \textit{Grubs} is a general concept and not necessary a pest (e.g., Wikipedia redirects it to \textit{Beetle}).
\end{itemize}
\section{Conclusions}
\label{sec:conclusions}

In this paper, we highlight the importance of solving the KGC task in real-world scenarios, i.e., considering the sparsity and interpretability issues.
We propose a novel model HoGRN, which advances the performance of existing embedding-based methods by improving the reasoning ability while keeping an efficient and explainable computation.
HoGRN has two main components, entity updating, and high-order reasoning, and is compatible with almost arbitrary scoring functions, which have been verified effective through extensive experiments. First, we evaluate HoGRN with various scoring functions on three sparse datasets. This presents excellent performance improvements. Second, we remove the component of high-order reasoning for ablation study, which demonstrates the value of capturing endogenous correlation to alleviate the sparsity issue. Third, we further analyze the model behavior at different levels of sparsity from two aspects. On the one hand, on a widely used full KGC dataset WN18RR, we achieve the best or second-best scores due to the fewer relations that benefit the component of high-order reasoning. On the other hand, HoGRN significantly outperforms different baselines in sparse settings and also performs comparably when rich information is available. Fourth, we explore the hyper-parameters, including the proposed relational masking ratio, for better optimization. Finally, case studies show examples of our explainable results.

In the future, we will attempt to explore the potential of HoGRN by considering different composition operations and integrating other scoring functions.
At the same time, injecting explicit rule information to assist the model training may also bring additional benefits.
Furthermore, it is also a promising direction to explore the application of HoGRN in real-life scenarios, such as recommendation systems and drug design.

% \newpage

\bibliographystyle{IEEEtran}
\bibliography{reference}
% \newpage

\end{document}